%% file: conference_101719.tex
\documentclass[conference]{IEEEtran}
\IEEEoverridecommandlockouts
\usepackage{cite}
\usepackage{amsmath,amssymb,amsfonts}
\usepackage{bm}
\usepackage{textcomp}
\usepackage{xcolor}
\usepackage{tikz}
\usetikzlibrary{shapes, arrows, positioning, calc, fit}
\usepackage{amsmath}
\usepackage[T1]{fontenc}
\usepackage{graphicx}
\usepackage{cite}
\usepackage{algpseudocode}
\usepackage{amsmath,amssymb,amsfonts}
\usepackage{amsthm}
\usepackage{textcomp}
\usepackage{xcolor}
\usepackage{subcaption}
\usepackage{upgreek}
\usepackage{booktabs}
\newcommand{\mycomment}[1]{}
\usepackage{balance}
\usepackage[hidelinks]{hyperref}
\def\BibTeX{{\rm B\kern-.05em{\sc i\kern-.025em b}\kern-.08em
    T\kern-.1667em\lower.7ex\hbox{E}\kern-.125emX}}

\makeatletter
\newcommand{\linebreakand}{%
  \end{@IEEEauthorhalign}
  \hfill\mbox{}\par
  \mbox{}\hfill\begin{@IEEEauthorhalign}
}

\usepackage[ruled,vlined,linesnumbered]{algorithm2e}

\begin{document}

    \title{Hess-MC$^2$: Sequential Monte Carlo Squared using Hessian Information and Second Order Proposals\\
\thanks{This research was funded in whole, or in part, by the Wellcome Trust [226691/Z/22/Z]. For the purpose of Open Access, the author has applied a CC BY public copyright license to any Author Accepted Manuscript version arising from this submission. CR was funded by the Wellcome CAMO-Net UK grant: 226691/Z/22/Z; JM and AM were funded by a Research Studentship jointly funded by the EPSRC Centre for Doctoral Training in Distributed Algorithms EP/S023445/1. This work was part-funded by Dstl in collaboration with the Royal Academy of Engineering via "Dstl-RAEng Research Chair in Information Fusion" under task RQ0000040616. SM thanks Dstl, UK MOD and the Royal Academy of Engineering for supporting this work. The views and conclusions contained in this paper are of the authors and should not be interpreted as representing the official policies, either expressed or implied, of the UK MOD or the UK Government. LD and PH were funded by EPSRC through the Big
Hypotheses under Grant EP/R018537/1\\
\bf{\footnotesize \textcopyright 2025 IEEE. Personal use of this material is permitted.
  Permission from IEEE must be obtained for all other uses, in any current or future
  media, including reprinting/republishing this material for advertising or promotional
  purposes, creating new collective works, for resale or redistribution to servers or
  lists, or reuse of any copyrighted component of this work in other works.}}
}

\author{\IEEEauthorblockN{Joshua Murphy\IEEEauthorrefmark{1}, Conor Rosato\IEEEauthorrefmark{2}, Andrew Millard\IEEEauthorrefmark{1},
   Lee Devlin\IEEEauthorrefmark{1},
   Paul Horridge\IEEEauthorrefmark{1} 
   and Simon Maskell\IEEEauthorrefmark{1}
}\\
\IEEEauthorblockA{\IEEEauthorrefmark{1} Department of Electrical Engineering and Electronics, University of Liverpool, United Kingdom\\
\IEEEauthorrefmark{2} Department of Pharmacology and Therapeutics, University of Liverpool, United Kingdom}

Email: \{joshua.murphy, cmrosa, andrew.millard, ljdevlin, p.horridge,  smaskell\}@liverpool.ac.uk}


\maketitle

\input{0_Abstract}

\input{1_Introduction}

\input{3_PF}

\input{5_SMCsquared}

\input{6_Results}

\input{7_Conclusions}

\input{8_References}


\end{document}

%% file: 0_Abstract.tex
\begin{abstract}

When performing Bayesian inference using Sequential Monte Carlo (SMC) methods, two considerations arise: the accuracy of the posterior approximation and computational efficiency. To address computational demands, Sequential Monte Carlo Squared (SMC$^2$) is well-suited for high-performance computing (HPC) environments. The design of the proposal distribution within SMC$^2$ can improve accuracy and exploration of the posterior as poor proposals may lead to high variance in importance weights and particle degeneracy. The Metropolis-Adjusted Langevin Algorithm (MALA) uses gradient information so that particles preferentially explore regions of higher probability. In this paper, we extend this idea by incorporating second-order information, specifically the Hessian of the log-target. While second-order proposals have been explored previously in particle Markov Chain Monte Carlo (p-MCMC) methods, we are the first to introduce them within the SMC$^2$ framework. Second-order proposals not only use the gradient (first-order derivative), but also the curvature (second-order derivative) of the target distribution. Experimental results on synthetic models highlight the benefits of our approach in terms of step-size selection and posterior approximation accuracy when compared to other proposals.

\end{abstract}

%% file: 1_Introduction.tex
\section{Introduction}
\label{sec:intro}

Bayesian inference offers a framework for uncertainty quantification, but remains computationally challenging in complex or high-dimensional models. Sequential Monte Carlo (SMC) methods are a flexible class of algorithms for approximating posterior distributions, especially in non-linear non-Gaussian problem settings. However, two key considerations must be balanced in SMC algorithms: computational efficiency and accuracy of the posterior approximation. Sequential Monte Carlo Squared (SMC$^2$) is one such method which is well-suited to estimating the dynamic states and parameters of state space models (SSM) \cite{chopin2013smc2}. SMC$^2$ consists of two layers: an inner particle filter (PF) \cite{arulampalam2002tutorial} layer which tracks the dynamic states and an SMC sampler \cite{del2006sequential} which estimates the parameters. Recent work in \cite{rosato2023log} introduced a framework for deploying SMC$^2$ on distributed memory architectures, which addresses the computational efficiency problem. This parallelizability represents a significant advantage over many other Bayesian inference algorithms. In \cite{rosato2023log}, only a random walk (RW) proposal was considered but RW can converge slowly in high-dimensional parameter spaces. To use gradient-based proposals the PF needs to be differentiated which can be problematic due to some of the discrete operations being non-differentiable \cite{chen2023overview}. Devising methods for differentiating PFs is therefore an active area of research \cite{nemeth2013particle,corenflos2021differentiable, karkus2018particle, rosato2022efficient}. In the context of parameter estimation for SSMs, first-order (FO) gradient information has been incorporated into particle-Markov Chain Monte Carlo (p-MCMC) methods \cite{particleLangevin1, nemeth2016particle}, which improved proposal efficiency. Second-order (SO) information which incorporates Hessian matrices has been included in general MCMC \cite{girolami2011riemann} and p-MCMC which has further improved inference quality \cite{Dahlin_2014}. Hessian matrices capture both the slope (via gradients) and the local curvature of the posterior, allowing for more informed and efficient proposals.

In the context of SMC$^2$, recent work to increase accuracy has focused on improving the proposal, including \cite{rosato2024enhanced}, in which a differentiable Common Random Number (CRN) PF \cite{rosato2022efficient} is employed to obtain the gradient of the log-likelihood with respect to the parameters. The resulting gradients were then used within the Metropolis-Adjusted Langevin Algorithm (MALA) proposal \cite{roberts1996exponential} to adaptively guide particles toward regions of higher posterior density using local information about the target. Building on this idea, we propose the use of SO information in the form of the Hessian matrix of the log-target density. Using automatic differentiation in PyTorch \cite{paszke2019pytorch}, we compute SO derivatives to construct curvature-aware proposals that improve the concentration and diversity of particles in high-probability regions. Choosing an appropriate step-size can be challenging for RW, FO, and SO proposals. Our analysis shows that accuracy is more sensitive to step-size in the case of RW proposals, exhibiting much higher variance compared to FO and SO methods.

The structure of this paper is as follows: in Section~\ref{sec:particle_filter} we describe a PF with the log-likelihood, FO and SO gradients outlined in Sections~\ref{sec:pf_likelihood}, \ref{sec:pf_gradient} and \ref{sec:second_pf_gradient}. The SMC sampler and different variants of the proposal are shown in Sections~\ref{sec:SMCsquared} and \ref{sec:proposals}, respectively. Two examples are shown in Section~\ref{sec:example} with conclusions and future work described in Section~\ref{sec:conclusions_future}.

%% file: 3_PF.tex
\section{Particle Filter}\label{sec:particle_filter}

State-Space Models (SSMs) can be used to represent the dependence of latent states in non-linear non-Gaussian dynamical systems. An SSM consists of a state equation and an observation equation parameterized by $\bm{\theta}$ 
\begin{align}\label{xt}
\mathbf{x}_{t} \mid \mathbf{x}_{t-1} &\sim p(\mathbf{x}_{t} \mid \mathbf{x}_{t-1}, \bm{\theta}),  \\ 
\mathbf{y}_{t} \mid \mathbf{x}_{t} &\sim p(\mathbf{y}_{t} \mid \mathbf{x}_{t},\bm{\theta}). 
\end{align}
An SSM sequentially simulates the latent states, $\mathbf{x}_{1:T}=\{\mathbf{x}_1,\dots, \mathbf{x}_t, \dots, \mathbf{x}_T\}$, conditional on data it receives at each time increment $t$, $\mathbf{y}_{1:T}=\{\mathbf{y}_1,\dots, \mathbf{y}_t, \dots, \mathbf{y}_T\}$, over $T$ timesteps. The joint distribution at timestep $t$ is 
\begin{align}
    p(\mathbf{x}_{1:t}, \mathbf{y}_{1:t}|\bm{\theta})=p(\mathbf{x}_1&|\bm{\theta})p(\mathbf{y}_1|\mathbf{x}_1,\bm{\theta})\times \nonumber\\
    &\prod_{\tau=2}^tp(\mathbf{x}_\tau|\mathbf{x}_{\tau-1}\bm{\theta})p(\mathbf{y}_\tau|\mathbf{x}_\tau,\bm{\theta}).\label{eq:joint}
\end{align}
%
A PF can be used to recursively approximate the distribution in \eqref{eq:joint} using a set of $N_x$ particles and an importance sampling approach. As time evolves, new particles are drawn from a proposal distribution, $q(\mathbf{x}_{t}|\mathbf{x}_{t-1}, \bm{\theta}, \mathbf{y}_{t})$. These particles are weighted at each timestep according to
\begin{align}
\mathbf{w}_{1:t}^{j}= \mathbf{w}_{1:t-1}^{j}\frac{p\left(\mathbf{y}_t|\mathbf{x}_t^{j},\bm{\theta}\right)p\left(\mathbf{x}_t^{j}|\mathbf{x}^{j}_{t-1},\bm{\theta}\right)}{q\left(\mathbf{x}_t^{j}|\mathbf{x}_{t-1}^{j}, \bm{\theta},\mathbf{y}_t\right)},\label{eq:pf_weightupdate}
\end{align}
%
%
Normalized weights $\Tilde{\mathbf{w}}^j_{1:t}$ are calculated by
\begin{equation}
    \tilde{\mathbf{w}}_{1:t}^{j} = \frac{\mathbf{w}_{1:t}^{j}}{\sum\nolimits_{j'=1}^{N_x} \mathbf{w}_{1:t}^{j'}}.
    \label{eq:normalise_pf}
\end{equation}
As time evolves, a small number of particles may have the majority of the weight, a phenomenon known as particle degeneracy. The number of effective samples can be used to monitor the performance of the samples is given by  
\begin{align} \label{eq: ess}
N_x^\text{eff} = \frac{1}{\sum\nolimits_{j=1}^{{N_x}} \left(\tilde{\mathbf{w}}_{t}^{j}\right)^2},
\end{align}
and we resample when $N_x^\text{eff}$ falls below ${{N_x}}/2$. Resampling involves sampling $N_x$ new samples with replacement from the existing set with probabilities according to their weights. Unbiased estimates of integrals of functions of the posterior with respect to \eqref{eq:joint} are realized as 
\begin{equation} 
        \int p(\mathbf{x}_{1:t},\mathbf{y}_{1:t}|\bm{\theta}) f(\mathbf{x}_{1:t})d\mathbf{x}_{1:t} = \frac{1}{N_x}\sum\nolimits_{j=1}^{N_x} \mathbf{w}^j_{1:t} f(\mathbf{x}^j_{t}). \label{eq:realised_estimates}
\end{equation}


\subsection{Log-Likelihood}\label{sec:pf_likelihood}

The log-posterior of the parameters (not the states) is 
        \begin{equation}
        \log \pi(\mathbf{\bm{\theta}})=\log p(\bm{\theta})+ \log p(\mathbf{y}_{1:T}|\bm{\theta}),\label{eq:log_posterior}
        \end{equation}
where $\log p(\bm{\theta})$ is the log-prior and $\log p(\mathbf{y}_{1:T}|\bm{\theta})$ represents the log-likelihood. An unbiased estimate of the log-likelihood is calculated from \eqref{eq:realised_estimates} with $f(\cdot)=1$.


\subsection{First-Order Gradients}\label{sec:pf_gradient}

In accordance with \eqref{eq:log_posterior}, the gradient of the log-posterior is 
\begin{align}
\nabla \log \pi(\mathbf{\bm{\theta}})=\nabla \log p(\mathbf{\bm{\theta}})+\nabla \log p(\mathbf{y}_{1:T}|\mathbf{\bm{\theta}}).
\label{eq:first_gradientlogposterior}
\end{align}
%
An approximation of the gradient of the likelihood is determined as a function of the derivative of the weights:
\begin{eqnarray}
	\frac{d}{d\bm{\theta}} {p(\mathbf{y}_{1:t}|\bm{\theta})} & \approx & \frac{1}{N_x} \sum_{j=1}^{N_x} \frac{d}{d\bm{\theta}}\mathbf{w}_{1:t}^{j}. \label{eq:dlikweightsum_grad}.
\end{eqnarray}
The details for deriving the gradient of the log-likelihood using the chain rule can be found in Appendix A of \cite{rosato2024enhanced}.

\subsection{Second-Order Gradients}\label{sec:second_pf_gradient}

In accordance with \eqref{eq:log_posterior}, the negative Hessian of the log-posterior is expressed as:
\begin{align}
\nabla^2 \log \pi(\mathbf{\bm{\theta}})=-\nabla^2 \log p(\mathbf{\bm{\theta}})-\nabla^2 \log p(\mathbf{y}_{1:T}|\mathbf{\bm{\theta}}).
\label{eq:second_gradientlogposterior}
\end{align}
%
An approximation to the negative Hessian of the log-likelihood is determined as a function of the derivative of the log-weights:
\begin{align}
	\frac{d^2}{d\bm{\theta}^2} {p(\mathbf{y}_{1:t}|\bm{\theta})} \approx & \frac{1}{N_x} \frac{d}{d\bm{\theta}}\Big( \sum_{j=1}^{N_x}\mathbf{\tilde{w}}_{1:t}^{j}\Big)\frac{d}{d\bm{\theta}}{\text{log}\mathbf{w}_{1:t}^{j}} \nonumber \\ &  + \frac{1}{N_x}\sum_{j=1}^{N_x}\mathbf{\tilde{w}}_{1:t}^{j}\frac{d^2}{d\bm{\theta}^2}{\text{log}\mathbf{w}_{1:t}^{j}}, 
 \label{eq:second_order}
\end{align}
where, by \eqref{eq:normalise_pf},
\begin{align}
\frac{d}{d\bm{\theta}}\sum_{j=1}^{N_x}\tilde{\mathbf{w}}_{1:t}^{j} & = \sum_{j=1}^{N_x} \frac{d}{d\bm{\theta}} \Big(\frac{\mathbf{w}_{1:t}^{j}}{\sum_{i=1}^{N_x} \mathbf{w}_{1:t}^j}\Big) \\ 
&  = \sum_{j=1}^{N_x}{\tilde{\mathbf{w}}_{1:t}^{j}\Big(\frac{d}{d\bm{\theta}}{\text{log}\mathbf{w}_{1:t}^{j}}}-{\frac{d}{d\bm{\theta}} {\log p(\mathbf{y}_{1:t}|\bm{\theta})\Big)}}.
\end{align}
Following the logic of \cite{rosato2024enhanced} Appendix A, we can find an estimate of the second derivative  of the log weights. Taking the second term in \eqref{eq:second_order},
\begin{align}
\frac{d^2}{d\bm{\theta}^2}{\text{log}\mathbf{w}_{1:t}^{j}}=\frac{d^2}{d^2\bm{\theta}}{\text{log}\mathbf{w}_{1:t-1}^{j}}+\frac{d^2}{d^2\bm{\theta}}{\text{log} p(\mathbf{y}_t|\mathbf{x}_{t}^{j})}.
\end{align}
Let 
\begin{align}
    	L\left(\mathbf{x}_t^{j}, \bm{\theta}, \mathbf{y}_t\right) &\triangleq \log p\left(\mathbf{y}_t | \mathbf{x}_{t}^{j}, \bm{\theta}\right), \label{eq:L}
\end{align}
where the likelihood is Gaussian with a variance that is independent of $x^{j}_t$, such that
\begin{align}
L\left(\mathbf{x}_t^{j}, \bm{\theta}, \mathbf{y}_t\right)&\triangleq \log\mathcal{N}\left(\mathbf{y}_t; h(x^{j}_t, \bm{\theta}), R(\bm{\theta})\right).
\end{align}
The SO derivative of the likelihood $L$ in \eqref{eq:L} is given by
\begin{align}
\frac{d^2}{d\bm{\theta}^2} L\left(\mathbf{x}_t^{j}, \bm{\theta}, \mathbf{y}_t\right) = & \frac{d}{d\bm{\theta}}\Big(\frac{\partial}{\partial h}\log\mathcal{N}(\mathbf{y}_t; h, R)\Big)\cdot\frac{dh}{d\bm{\theta}} \nonumber \\
&+ \frac{\partial}{\partial h}\log\mathcal{N}(\mathbf{y}_t; h, R)\cdot \frac{d^2h}{d\bm{\theta}^2} \nonumber \\
&+ \frac{d}{d\bm{\theta}} \Big(\frac{\partial}{\partial R}\log\mathcal{N}(\mathbf{y}_t; h, R)\Big)\cdot\frac{d R}{d\bm{\theta}} \nonumber \\
&+ \frac{d}{d R} \log\mathcal{N}(\mathbf{y}_t; h, R)\cdot\frac{d^2R}{d^2 \bm{\theta}},
\end{align}
for
\begin{equation}
    \frac{dh}{d\bm{\theta}} = \frac{\partial h}{\partial \mathbf{x}^{j}_t}\frac{d \mathbf{x}^{j}_t}{d \bm{\theta}}  + \frac{\partial h}{\partial \bm{\theta}},
\end{equation}
Then,
\begin{align}
\frac{d}{d\bm{\theta}}\Big(\frac{\partial}{\partial h}\log\mathcal{N}(\mathbf{y}_t; h, R)\Big) = & \frac{\partial^2}{\partial^2 h}\log\mathcal{N}(\mathbf{y}_t; h, R) \cdot \frac{dh}{d\bm{\theta}} \nonumber \\
&+ \frac{\partial^2}{\partial h \partial R} \log\mathcal{N}(\mathbf{y}_t; h, R) \cdot \frac{d R}{d \bm{\theta}},
\end{align}
\begin{align}
\frac{dh^2}{d\bm{\theta}^2} = & \frac{\partial^2h}{\partial\mathbf{x}^{j2}_t}\Big(\frac{d\mathbf{x}^{j}_t}{d \bm{\theta}}\Big)^2+\frac{2\partial^2 h}{\partial \bm{\theta} \partial \mathbf{x}^{j}_t}\Big(\frac{d\mathbf{x}^{j}_t}{d \bm{\theta}}\Big)+\frac{\partial h}{\partial  \mathbf{x}^{j}_t}\Big(\frac{d^2\mathbf{x}^{j}_t}{d \bm{\theta}^2}\Big) \nonumber \\
&+\frac{\partial^2h}{\partial\bm{\theta}}
\end{align}
and
\begin{align}
\frac{d}{d\bm{\theta}}\Big(\frac{\partial}{\partial R}\log\mathcal{N}(\mathbf{y}_t; h, R)\Big) = & \frac{\partial^2}{\partial h \partial R}\log\mathcal{N}(\mathbf{y}_t; h, R)\cdot \frac{\partial h}{\partial\bm{\theta}} \nonumber \\
& +  \frac{d^2}{d^2 R} \log\mathcal{N}(\mathbf{y}_t; h, R) \cdot \frac{d R}{d\bm{\theta}}.
\end{align}

%% file: 5_SMCsquared.tex
\section{Sequential Monte Carlo Squared}
\label{sec:SMCsquared}
 SMC$^2$ runs for K iterations, targeting the posterior of the parameters $\pi(\mathbf{\bm{\theta}})$ at each iteration $k$. The joint distribution from all states until $k=K$ is defined to be
    \begin{equation}
        \pi(\mathbf{\bm{\theta}}_{1:K}) = \pi(\mathbf{\bm{\theta}}_{K}) \prod_{k=2}^{K} L_k(\mathbf{\bm{\theta}}_{k-1} | \mathbf{\bm{\theta}}_{k}),
    \end{equation}
where $L_k(\mathbf{\bm{\theta}}_{k-1} | \mathbf{\bm{\theta}}_{k})$ is the L-kernel, a user-defined probability distribution. At $k=1$, $N$ samples are drawn from a prior distribution $q_1(\cdot)$ and weighted according to
\begin{equation} 
    \mathbf{v}^i_1 = \frac{\pi(\mathbf{\bm{\theta}}^i_1)}{q_1(\mathbf{\bm{\theta}}^i_1)}\label{init_weights}.
\end{equation}

At $k>1$, subsequent samples are proposed based on samples from the previous iteration via a proposal distribution, $q(\mathbf{\bm{\theta}}^i_k|\mathbf{\bm{\theta}}^i_{k-1})$. These samples are weighted according to 
\begin{equation}
    \mathbf{v}^i_{k} = \mathbf{v}^i_{k-1} \frac{\pi(\mathbf{\bm{\theta}}^i_{k})}{\pi(\mathbf{\bm{\theta}}^i_{k-1})} \frac{L_k(\mathbf{\bm{\theta}}^i_{k-1}|\mathbf{\bm{\theta}}^i_{k})}{q_k(\mathbf{\bm{\theta}}^i_{k}|\mathbf{\bm{\theta}}^i_{k-1})}.\label{eq:l_weights}
\end{equation}
At the SMC sampler level we employ a parallelized version of systematic resampling \cite{Alessandro5} when the effective sample size, calculated as in \eqref{eq: ess}, goes below $N/2$. Estimates of functions on the distribution are realized by
\begin{equation}
    \mathbb{E}_{\pi}\left[ f(\bm{\theta})\right] \approx \sum_{i=1}^{N} \Tilde{\mathbf{v}}^i_{k} f(\bm{\theta}^i_{k}), \label{eq:realised_estimates_smc}
\end{equation}
with samples from previous iterations incorporated through recycling \cite{nguyen2015efficient}.

\subsection{Proposals}
\label{sec:proposals}
In this paper we consider RW, FO, and SO proposals and derive L-kernels based off a change of variables (CoV). The proposals are 
\begin{align}
     q_k(\bm{\theta}_{k}|\bm{\theta}_{k-1}) = \begin{cases}
         \mathcal{N}(\bm{\theta}_{k-1}, \Gamma) &\text{RW} \\
         \mathcal{N}(\bm{\theta}_{k-1}+\frac{\Gamma}{2}G_{k-1}, \Gamma) &\text{FO} \\
         \mathcal{N}(\bm{\theta}_{k-1}+\frac{\Gamma}{2}H_{k-1}G_{k-1},\Gamma H_{k-1}), &\text{SO}
     \end{cases} \label{eq:dahlin}
\end{align}
where we let the gradient and Hessian of the log-posterior be $G_{k-1}=\nabla\log\pi(\bm{\theta}_{k-1})$ and $H_{k-1}=(-\nabla^2\log\pi(\bm{\theta}_{k-1}))^{-1}$. These proposals are parameterized in terms of a scalar step size $\epsilon$ such that $\Gamma=\epsilon^2\mathbf{I}$.

\subsubsection{Random-Walk}
\label{sec:rw_proposals}
\noindent The proposal and L-kernel are evaluated in \eqref{eq:l_weights} as
\begin{align}
    q_k(\bm{\theta}_{k}|\bm{\theta}_{k-1})=\mathcal{N}(\bm{\theta}_{k}; \bm{\theta}_{k-1},\epsilon^2\mathbf{I}),\\
    L_k(\bm{\theta}_{k-1}|\bm{\theta}_k)=\mathcal{N}(\bm{\theta}_{k-1}; \bm{\theta}_k,\epsilon^2\mathbf{I}).
\end{align}

\subsubsection{First-Order}
\label{sec:fo_proposals}
We can rewrite the FO proposal in \eqref{eq:dahlin} as
\begin{align}
    \bm{\theta}_k = \bm{\theta}_{k-1} + \frac{\epsilon^2}{2}G_{k-1} +\epsilon\mathbf{p}_{k-1}, \label{eq:1st_full}
\end{align}

\noindent and can then propose samples according to
\begin{align}
    \mathbf{p}_{k-1} &\sim \mathcal{N}(0, \mathbf{I}), \\
    \mathbf{p}_{k-0.5} &= \frac{\epsilon}{2}G_{k-1} + \mathbf{p}_{k-1},\\
    \bm{\theta}_k &= \bm{\theta}_{k-1} +\epsilon\mathbf{p}_{k-0.5}, \\
    \mathbf{p}_{k} &= \frac{\epsilon}{2}G_{k} + \mathbf{p}_{k-0.5},
\end{align}
which resembles the leapfrog integrator. As in \cite{devlin2024no} leapfrog is considered to be a function, $f_{LF}$ which transforms $\bm{\theta}_{k-1}$ to $\bm{\theta}_{k}$ so the proposal can be rewritten through a CoV
\begin{align}
     q_k(\bm{\theta}_{k}|\bm{\theta}_{k-1}) &= q_k^p(f_{LF}(\bm{\theta}_{k-1}, \mathbf{p}_{k-1})|\bm{\theta}_{k-1}),\\
     &= q_k^p(\mathbf{p}_{k-1}|\bm{\theta}_{k-1})\begin{vmatrix}\frac{df_{LF}(\bm{\theta}_{k-1}, \mathbf{p}_{k-1})}{d\mathbf{p}_{k-1}}\end{vmatrix}^{-1} \\
    &=  \mathcal{N}( \mathbf{p}_{k-1}; 0, \mathbf{I})\begin{vmatrix}\frac{df_{LF}(\bm{\theta}_{k-1}, \mathbf{p}_{k-1})}{d\mathbf{p}_{k-1}}\end{vmatrix}^{-1}.
\end{align}
The L-kernel can likewise be written in terms of $f_{LF}$ as leapfrog is reversible so that the negative of the final momentum $\mathbf{p}_{k}$ will bring a sample from $\bm{\theta}_k$ to $\bm{\theta}_{k-1}$
\begin{align}
    L_k(\bm{\theta}_{k-1}|\bm{\theta}_k)&= L_k^p(-\mathbf{p}_{k}|\bm{\theta}_k)\begin{vmatrix}\frac{df_{LF}(\bm{\theta}_k, -\mathbf{p}_{k})}{d\mathbf{p}_{k}}\end{vmatrix}^{-1} \nonumber\\
    &=  \mathcal{N}(-\mathbf{p}_{k}; 0, \mathbf{I})\begin{vmatrix}\frac{df_{LF}(\bm{\theta}_k, -\mathbf{p}_{k})}{d\mathbf{p}_{k}}\end{vmatrix}^{-1}. \label{eq:1st_l_cov}
\end{align}

\subsubsection{Second-Order}
\label{sec:so_proposals}

As in Section \ref{sec:fo_proposals}, we write the SO proposal in leapfrog form
\begin{align}
    \mathbf{p}_{k-1} &\sim \mathcal{N}(0, H_{k-1}^{-1}), \\
    \mathbf{p}_{k-0.5} &= \frac{\epsilon}{2}G_{k-1} + \mathbf{p}_{k-1},\\
    \bm{\theta}_k &= \bm{\theta}_{k-1} +\epsilon H_{k-1}\mathbf{p}_{k-0.5},\\
    \mathbf{p}_{k} &= \frac{\epsilon}{2}G_{k} + \mathbf{p}_{k-0.5},
\end{align}
and through a CoV can evaluate the proposal and L-kernel as
\begin{align}
    q_k(\bm{\theta}_{k}|\bm{\theta}_{k-1}) &=  \mathcal{N}( \mathbf{p}_{k-1}; 0, H_{k-1}^{-1})\begin{vmatrix}\frac{df_{LF}(\bm{\theta}_{k-1}, \mathbf{p}_{k-1})}{d\mathbf{p}_{k-1}}\end{vmatrix}^{-1}, \\ 
    L_k(\bm{\theta}_{k-1}|\bm{\theta}_k)&=\mathcal{N}(-\mathbf{p}_{k}; 0, H_{k-1}^{-1})\begin{vmatrix}\frac{df_{LF}(\bm{\theta}_k, -\mathbf{p}_{k})}{d\mathbf{p}_{k}}\end{vmatrix}^{-1}.\label{eq:2nd_q_cov}
\end{align}
There are cases where $H_{k-1}$ is not a positive semi-definite matrix which is likely to occur far from posterior modes where the amount of information is limited. While there are heuristics to handle non positive semi-definite estimates of the Hessian \cite{nocedal1999numerical}, we choose to revert to the FO so as not to rely on insufficient information to effectively inform moves in the parameter space.

%% file: 6_Results.tex
\section{Examples}\label{sec:example}
For both examples considered, we generate data for 50 different seeds and perform parameter inference for each dataset and average the results. The algorithmic setup consists of using $N_x=500$ particles in each PF and $N=32$ samples in the SMC sampler over $K=15$ iterations. We use the parallelized SMC$^2$ framework outlined in \cite{rosato2022efficient, rosato2024enhanced} which enables multiple instances of the computationally intensive inner PF to be run in parallel. For each proposal we take 20 step sizes in a range that maintains numerical stability. We evaluate performance in terms of root mean square error (RMSE) between the true and the mean parameter estimates obtained from the SMC sampler. The relative runtime (RR) of FO and SO proposals are calculated with respect to RW. The analysis was performed on a distributed
memory cluster equipped with two Xeon Gold 6138 CPUs,
providing 384GB of memory and 40 cores. The code can be found here \footnote{\href{https://github.com/j-j-murphy/SMC-Squared-Langevin}{https://github.com/j-j-murphy/SMC-Squared-Langevin}}.

\subsection{Linear Gaussian State Space Model}\label{sec:LGSSM}
We consider a Linear Gaussian State Space (LGSS) model outlined in \cite{dahlin2019getting, rosato2022efficient} with state and observation equations 
\begin{align}
\mathbf{x}_{t} \mid \mathbf{x}_{t-1} &\sim \mathcal{N}\left(\mathbf{x}_{t} ; \mu \mathbf{x}_{t-1}, \phi^{2}\right),  \\\mathbf{y}_{t} \mid \mathbf{x}_{t} &\sim \mathcal{N}\left(\mathbf{y}_{t} ; \mathbf{x}_{t}, \sigma^{2}\right), 
\label{LGSSyt}
\end{align}
where $\bm{\theta}=\left\{\mu, \phi, \sigma \right\}=\{0.75,1,1\}$. The PF uses the ``optimal'' proposal which can be derived from \eqref{LGSSyt}
\begin{equation}
q\left(\mathbf{x}_{t}|\mathbf{x}_{t-1}, \bm{\theta},\mathbf{y}_t\right) =\mathcal{N}\left(\mathbf{x}_{t} ; \rho^{2}\left[\sigma^{-2} \mathbf{y}_{t}+\phi^{-2} \mu \mathbf{x}_{t-1}\right], \rho^{2}\right),
\end{equation}
with $\rho^{-2}=\phi^{-2}+\sigma^{-2}$. The PF weights are updated using
\begin{equation}
\mathbf{w}_{t}^{j}=\mathbf{w}_{t-1}^{j}\mathcal{N}\left(\mathbf{y}_t; \mu \mathbf{x}_{t}, \rho^2\right).
\end{equation}
Each random seed is run for $T=500$. We use $\mathcal{U}(0,1)$, $\mathcal{U}(0,2)$ and $\mathcal{U}(0,2)$ as priors over the parameters. Figure \ref{fig:rmse_box} demonstrates how there is far less variability in RMSE across step size when using FO and SO compared to RW. Table \ref{Table:LGSSM_results} shows a significant benefit in the RMSE when using gradient-based proposals. While runtime for both FO and SO is higher, this is mitigated by less need for manual step-size tuning. We note that none of the 20 RW step-sizes leads to a better RMSE than SO. In the context of this model, the benefit of SO relative to FO is arguably insufficient to warrant the relatively large increase in runtime.
\begin{table}[ht]
\caption{LGSS Results for median $\epsilon$ by RMSE.}
\renewcommand{\arraystretch}{1.2}
\centering
\begin{tabular}{cccccc}
\hline 
\hline 
\textbf{Prop.-$\epsilon$} & $\mathbf{E[\mu]}$ & $\mathbf{E[\phi]}$ & $\mathbf{E[\sigma]}$ & \textbf{RMSE} & \textbf{RR}\\
\hline 
RW-0.7 & $0.606$ & $1.22$ & $0.926$ & $56.9 \times 10^{-3}$&$1.00$ \\
FO-0.03 & $0.717$ & $1.06$ & $0.958$ & $1.93 \times 10^{-3}$&$2.45$\\
SO-1.55 & $0.734$ & $1.04$ & $0.955$ & $1.50 \times 10^{-3}$&$12.2$\\
\hline \hline
\end{tabular}
\label{Table:LGSSM_results}
\end{table}

\subsection{Susceptible-Infected-Recovered Model}
\begin{figure}[!htb]
\begin{minipage}[b]{.48\linewidth}
  \centering
  \centerline{\includegraphics[width=4.0cm]{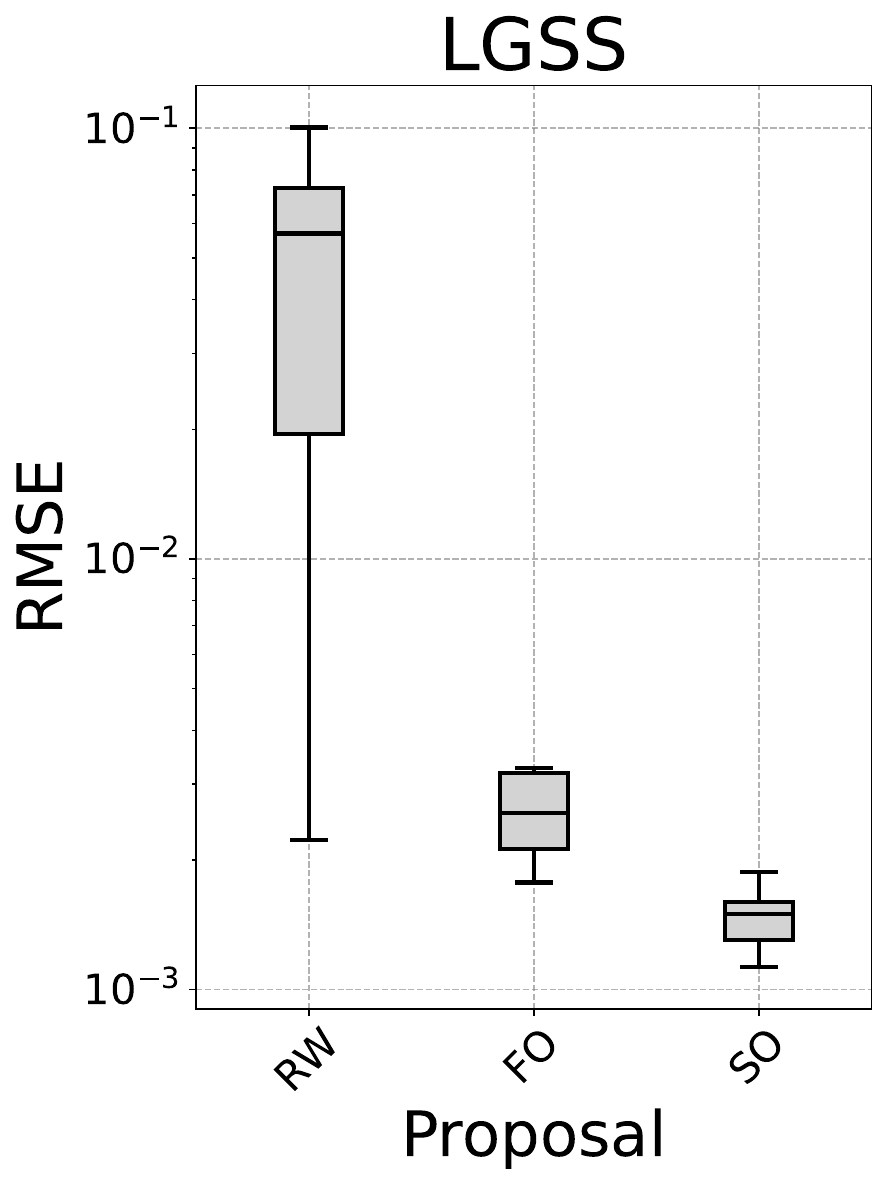}}
\end{minipage}
\hfill
\begin{minipage}[b]{0.48\linewidth}
  \centering
  \centerline{\includegraphics[width=4.0cm]{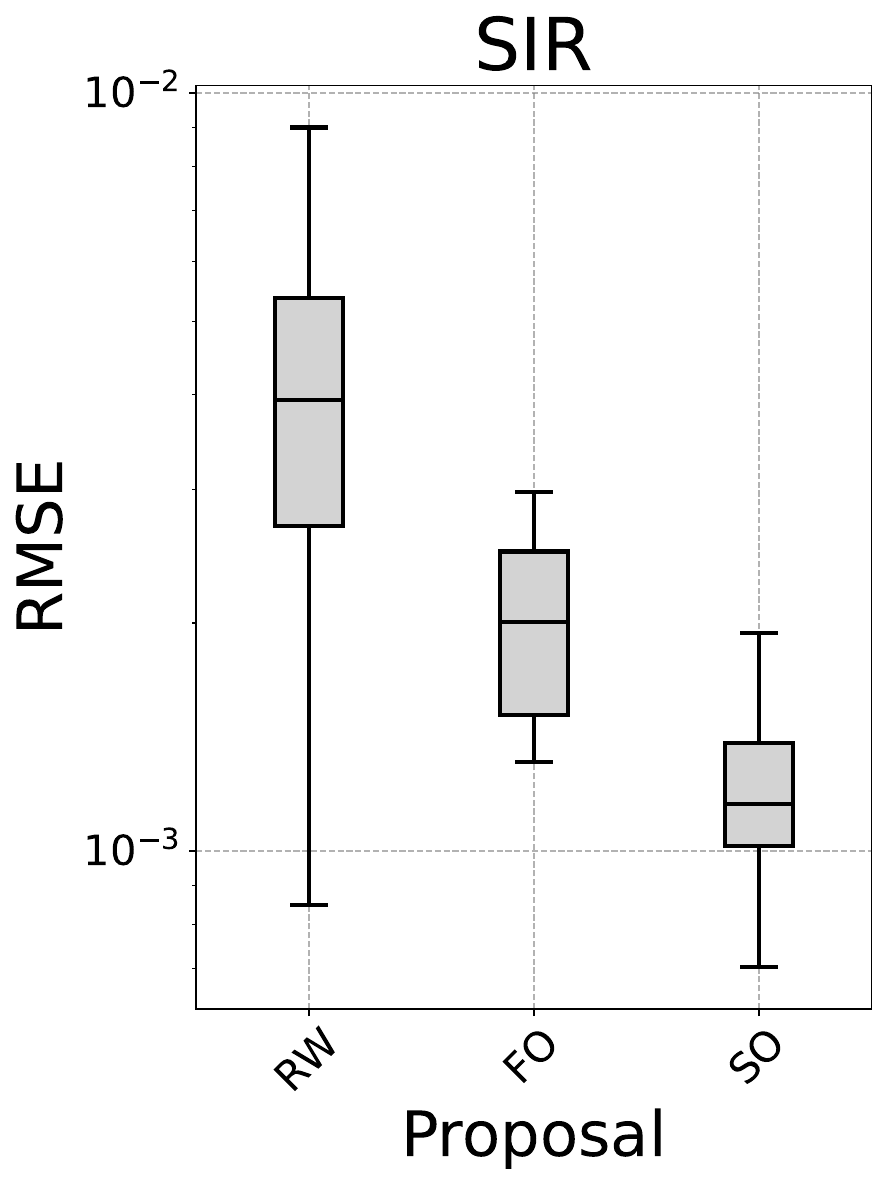}}
\end{minipage}
\caption{RMSE distribution over 20 step-sizes by proposal.}
\label{fig:rmse_box}
\end{figure}

 We also consider the Susceptible-Infected-Recovered (SIR) epidemiological model outlined in \cite{sheinson2014comparison,rosato2022inference}. A discrete time approximation of the SIR model is presented below
\begin{align}
&\mathbf{S}_{t}=\mathbf{S}_{t-1}-\beta \mathbf{I}_{t-1} \mathbf{S}_{t-1}+\epsilon_{\beta}, \label{SIRR:s}\\
& \mathbf{I}_{t}= \mathbf{I}_{t-1}+\beta  \mathbf{I}_{t-1} \mathbf{S}_{t-1}-\gamma \label{SIRR:i}  \mathbf{I}_{t-1}-\epsilon_{\beta}+\epsilon_{\gamma},\\
& \mathbf{R}_{t}=N_{pop} - \mathbf{S}_{t} - \mathbf{I}_{t} \label{SIRR:r},
\end{align}
where $\bm{\theta}=\{\beta, \gamma\}$. The number of individuals in each compartment at time $t=0$ are denoted $\mathbf{S}_{0}=N_{pop}-\mathbf{I}_{0}$, $\mathbf{I}_{0}=1$ and $\mathbf{R}_{0}=0$ and the total population is $N_{pop}=763$. Stochasticity is introduced by including a noise term, $\epsilon_{\bm{\theta}}$ for each parameter, which is independently drawn from $\mathcal{N}(0, 0.5)$. PF weights are updated using
\begin{equation}
\mathbf{w}_{t}^{j}=\mathbf{w}_{t-1}^{j}\mathcal{P}(\mathbf{y}_t; \mathbf{I}_t).
\end{equation}  
Each random seed is run for $T=36$ and we use $\mathcal{U}(0,1)$ as priors for both parameters. The results reflect those of the LGSS model. In Fig. \ref{fig:rmse_box}, there is more variance in the RMSE estimates with RW compared to FO and more of a spread in FO than SO. However, as is also evident in Fig. \ref{fig:rmse_box}, when looking at the best performing $\epsilon$ of RW, it performs better than that of FO. This demonstrates that it is possible to select an $\epsilon$ that makes RW competitive with gradient-based proposals but time consuming as it required evaluating 20 different step-sizes. On this smaller problem, the difference in RR and median RMSE is less pronounced for FO and SO over RW and the vast majority of the 20 RW step-sizes do not produce a better RMSE when compared to SO. In this model, the increase in runtime appears to offer commensurate improvements in RMSE to that which would be likely to be achieved using a correspondingly larger number of samples.

\begin{table}[ht]
    \centering
    \caption{SIR Results for median $\epsilon$ by RMSE.}
    \renewcommand{\arraystretch}{1.2}
    \begin{tabular}{ccccc}
    \hline 
    \hline 
    \textbf{Prop.-$\epsilon$} & $\mathbf{E[\beta]}$ & $\mathbf{E[\gamma]}$ & \textbf{RMSE} & \textbf{RR}\\
    \hline 
    RW-0.55 & $0.605$ & $0.299$ & $3.94 \times 10^{-3}$  & $1.00$\\
    FO-0.008& $0.580$ & $0.292$ & $2.01 \times 10^{-3}$ & $1.52$\\
    SO-2.05& $0.592$ & $0.295$ & $1.15 \times 10^{-3}$ & 5.26\\
    \hline \hline
    \end{tabular}
    \label{Table:SIR_result}
\end{table}

%% file: 7_Conclusions.tex
\section{Conclusions and Future Work}\label{sec:conclusions_future}

In this paper, we extend the work of \cite{rosato2024enhanced} by computing SO gradients from a CRN-PF and incorporating them into SO proposals in the form of a Hessian matrix. We document how selecting an appropriate step-size can be time-consuming and show that SO, when compared to FO and RW, has less variability in the RMSE estimates of the parameters of an SSM. The two chosen examples are relatively simple, low-dimensional models and further work may chose to explore whether the benefits of SO proposals extend to more complex  high-dimensional models. 

One recent method of adapting the step-size in SMC samplers can be found in \cite{kim2025tuning} which could be applicable in this work. Additional gradient-based proposals that could be employed include HMC, ChEES \cite{millard2025incorporating} and NUTS which was first described in \cite{rosato2022efficient} for particle MCMC.

%% file: 8_References.tex
\bibliographystyle{ieeetr}
\bibliography{bibliography.bib}

%% file: conference_101719.bbl
\begin{thebibliography}{10}

\bibitem{chopin2013smc2}
N.~Chopin, P.~E. Jacob, and O.~Papaspiliopoulos, ``{SMC}$^2$: an efficient algorithm for sequential analysis of state space models,'' {\em Journal of the Royal Statistical Society Series B: Statistical Methodology}, vol.~75, no.~3, pp.~397--426, 2013.

\bibitem{arulampalam2002tutorial}
M.~S. Arulampalam, S.~Maskell, N.~Gordon, and T.~Clapp, ``A tutorial on particle filters for online nonlinear/non-gaussian bayesian tracking,'' {\em IEEE Transactions on signal processing}, vol.~50, no.~2, pp.~174--188, 2002.

\bibitem{del2006sequential}
P.~Del~Moral, A.~Doucet, and A.~Jasra, ``Sequential {m}onte {c}arlo samplers,'' {\em Journal of the Royal Statistical Society Series B: Statistical Methodology}, vol.~68, no.~3, pp.~411--436, 2006.

\bibitem{rosato2023log}
C.~Rosato, A.~Varsi, J.~Murphy, and S.~Maskell, ``An {O}(log$_2${N}) {SMC}$^2$ algorithm on distributed memory with an approx. optimal l-kernel,'' in {\em 2023 IEEE Symposium Sensor Data Fusion and International Conference on Multisensor Fusion and Integration (SDF-MFI)}, pp.~1--8, IEEE, 2023.

\bibitem{chen2023overview}
X.~Chen and Y.~Li, ``An overview of differentiable particle filters for data-adaptive sequential {B}ayesian inference,'' {\em Foundations of Data Science}, pp.~0--0, 2023.

\bibitem{nemeth2013particle}
C.~Nemeth, P.~Fearnhead, and L.~Mihaylova, ``Particle approximations of the score and observed information matrix for parameter estimation in state--space models with linear computational cost,'' {\em Journal of Computational and Graphical Statistics}, vol.~25, no.~4, pp.~1138--1157, 2016.

\bibitem{corenflos2021differentiable}
A.~Corenflos, J.~Thornton, G.~Deligiannidis, and A.~Doucet, ``Differentiable particle filtering via entropy-regularized optimal transport,'' in {\em International Conference on Machine Learning}, pp.~2100--2111, PMLR, 2021.

\bibitem{karkus2018particle}
P.~Karkus, D.~Hsu, and W.~S. Lee, ``Particle filter networks with application to visual localization,'' in {\em Conference on robot learning}, pp.~169--178, PMLR, 2018.

\bibitem{rosato2022efficient}
C.~Rosato, L.~Devlin, V.~Beraud, P.~Horridge, T.~B. Sch{\"o}n, and S.~Maskell, ``Efficient learning of the parameters of non-linear models using differentiable resampling in particle filters,'' {\em IEEE Transactions on Signal Processing}, vol.~70, pp.~3676--3692, 2022.

\bibitem{particleLangevin1}
J.~Dahlin, F.~Lindsten, and T.~B. Sch{\"o}n, ``Particle metropolis hastings using {L}angevin dynamics,'' in {\em 2013 IEEE International Conference on Acoustics, Speech and Signal Processing}, pp.~6308--6312, IEEE, 2013.

\bibitem{nemeth2016particle}
C.~Nemeth, C.~Sherlock, and P.~Fearnhead, ``Particle metropolis-adjusted {L}angevin algorithms,'' {\em Biometrika}, vol.~103, no.~3, pp.~701--717, 2016.

\bibitem{girolami2011riemann}
M.~Girolami and B.~Calderhead, ``Riemann manifold {L}angevin and {H}amiltonian monte carlo methods,'' {\em Journal of the Royal Statistical Society Series B: Statistical Methodology}, vol.~73, no.~2, pp.~123--214, 2011.

\bibitem{Dahlin_2014}
J.~Dahlin, F.~Lindsten, and T.~B. Sch{\"o}n, ``Particle metropolis--hastings using gradient and {H}essian information,'' {\em Statistics and computing}, vol.~25, no.~1, pp.~81--92, 2015.

\bibitem{rosato2024enhanced}
C.~Rosato, J.~Murphy, A.~Varsi, P.~Horridge, and S.~Maskell, ``Enhanced {SMC}$^2$: Leveraging gradient information from differentiable particle filters within {L}angevin proposals,'' in {\em 2024 IEEE International Conference on Multisensor Fusion and Integration for Intelligent Systems (MFI)}, pp.~1--8, IEEE, 2024.

\bibitem{roberts1996exponential}
G.~O. Roberts and R.~L. Tweedie, ``Exponential convergence of {L}angevin distributions and their discrete approximations,'' {\em Bernoulli}, pp.~341--363, 1996.

\bibitem{paszke2019pytorch}
A.~Paszke, ``Pytorch: An imperative style, high-performance deep learning library,'' {\em arXiv preprint arXiv:1912.01703}, 2019.

\bibitem{Alessandro5}
A.~Varsi, S.~Maskell, and P.~G. Spirakis, ``An {O}(log$_2${N}) fully-balanced resampling algorithm for particle filters on distributed memory architectures,'' {\em Algorithms}, vol.~14, no.~12, pp.~342--362, 2021.

\bibitem{nguyen2015efficient}
T.~L.~T. Nguyen, F.~Septier, G.~W. Peters, and Y.~Delignon, ``Efficient sequential {M}onte-{C}arlo samplers for {B}ayesian inference,'' {\em IEEE Transactions on Signal Processing}, vol.~64, no.~5, pp.~1305--1319, 2015.

\bibitem{devlin2024no}
L.~Devlin, M.~Carter, P.~Horridge, P.~L. Green, and S.~Maskell, ``The no-u-turn sampler as a proposal distribution in a sequential {M}onte {C}arlo sampler without accept/reject,'' {\em IEEE Signal Processing Letters}, 2024.

\bibitem{nocedal1999numerical}
J.~Nocedal and S.~J. Wright, {\em Numerical optimization}.
\newblock Springer, 1999.

\bibitem{dahlin2019getting}
J.~Dahlin and T.~B. Sch{\"o}n, ``Getting started with particle metropolis-hastings for inference in nonlinear dynamical models,'' {\em Journal of Statistical Software}, vol.~88, pp.~1--41, 2019.

\bibitem{sheinson2014comparison}
D.~M. Sheinson, J.~Niemi, and W.~Meiring, ``Comparison of the performance of particle filter algorithms applied to tracking of a disease epidemic,'' {\em Mathematical biosciences}, vol.~255, pp.~21--32, 2014.

\bibitem{rosato2022inference}
C.~Rosato, J.~Harris, J.~Panovska-Griffiths, and S.~Maskell, ``Inference of stochastic disease transmission models using particle-{MCMC} and a gradient based proposal,'' in {\em 2022 25th International Conference on Information Fusion (FUSION)}, pp.~1--8, IEEE, 2022.

\bibitem{kim2025tuning}
K.~Kim, Z.~Xu, J.~R. Gardner, and T.~Campbell, ``Tuning sequential {M}onte {C}arlo samplers via greedy incremental divergence minimization,'' {\em arXiv preprint arXiv:2503.15704}, 2025.

\bibitem{millard2025incorporating}
A.~Millard, J.~Murphy, D.~Frisch, and S.~Maskell, ``Incorporating the chees criterion into sequential monte carlo samplers,'' {\em arXiv preprint arXiv:2504.02627}, 2025.

\end{thebibliography}
